\documentclass[10pt]{article}
\usepackage{cogsci}
\usepackage{apacite}
\usepackage{times}
\usepackage{mathptmx}
\usepackage{pst-node,pst-plot,pst-grad}
\usepackage{amsmath,amssymb}
\usepackage{latexsym}
\usepackage{psfrag}

\AtBeginDocument{\setlength\baselineskip{11.55pt}}

\def\citeS#1{\citeauthor{#1}'s \citeyear{#1}}

\def\bmath#1{\textbf{\itshape #1}}

\title{A Neurobiologically Motivated Analysis of Distributional Semantic Models}

\author{{\large\bf Akira Utsumi (utsumi@uec.ac.jp)} \\
  Department of Informatics, The University of Electro-Communications\\
  1-5-1, Chofugaoka, Chofushi, Tokyo 182-8585, Japan}

\begin{document}

\newcommand{\BIBPATH}{../bib}
\setcounter{totalnumber}{5}
\setcounter{topnumber}{4}
\renewcommand{\dbltopfraction}{1}
\renewcommand{\topfraction}{1}
\renewcommand{\textfraction}{0}

\maketitle

\begin{abstract}
  The pervasive use of distributional semantic models or word embeddings
  in a variety of research fields is due to their remarkable ability 
  to represent the meanings of words for both practical application
  and cognitive modeling.
  However, little has been known about what kind of information is encoded
  in text-based word vectors.
  This lack of understanding is particularly problematic 
  when word vectors are regarded as a model of
  semantic representation for abstract concepts.
  This paper attempts to reveal the internal 
  information of distributional word vectors by the analysis
  using \citeS{Binder16} brain-based vectors, 
  explicitly structured conceptual representations
  based on neurobiologically motivated attributes.
  In the analysis, 
  the mapping from text-based vectors to brain-based vectors is trained and
  prediction performance is evaluated by comparing the 
  estimated and original brain-based vectors.
  The analysis demonstrates that
  social and cognitive information is better encoded in text-based word vectors,
  but emotional information is not.
  This result is discussed in terms of embodied theories for abstract concepts.

  \vspace*{1mm}
  \textbf{Keywords:}
  Distributional semantic models; Word vectors; Brain-based representation;
  Embodied cognition; Emotional and social information; Abstract concepts
\end{abstract}

\section{Introduction}
One of the most important advances in the study of semantic processing
is the development of distributional semantic models for representing
word meanings.
In the distributional semantic model,
words are represented as high-dimensional vectors, which 
can be learned from the distributional statistics of word occurrence
in large collections of text.
Any words that occur in the corpus can be learned 
regardless of their part-of-speech class, abstractness, novelty and familiarity.
This is an important advantage of text-based distributional semantic models
over other spatial models of semantic representation such as
feature-based vectors \cite{Andrews09} and image-based vectors \cite{Silberer17}.

Word vectors have been employed in a variety of research fields and
many successful results have been obtained.
In the field of natural language processing (NLP), 
deep learning has recently been applied to 
a number of NLP tasks such as machine translation and automatic
summarization, and achieved the impressive performance
as compared to the traditional statistical methods.
One of the reasons for the successful results is 
the use of word vectors as semantic representations
for the input and output of recurrent neural networks \cite{Goldberg17}.
Research on cognitive science also benefits greatly from distributional
semantic models \cite{Jones15}.
Word vectors have been demonstrated to explain a number of cognitive phenomena 
relevant to semantic memory or mental lexicon, such as
word association \cite{Jones17,Utsumi:15:paper}, semantic priming \cite{Mandera17}, 
semantic transparency \cite{Marelli15} and conceptual combination \cite{Vecchi17}.
Furthermore, recent brain imaging studies have demonstrated that 
distributional word vectors have a powerful ability to predict the neural brain activity
evoked by lexical processing \cite{Mitchell08,Huth16,Guclu15}.
These voxel-wise modeling by word vectors is expected to open a door
for brain-machine interfaces.

Despite the fact that successful results are obtained in many research fields,
little has been known about 
what kind of information or knowledge is encoded in word vectors.
This lack of understanding makes distributional semantic models
unable to predict human language behavior and performance 
at the same level of detail and precision of other cognitive models.
It also limits further improvements on the practical performance 
of word vectors for many NLP tasks.

In this paper, therefore, we attempt to reveal the internal information
(or knowledge) encoded in text-based word vectors generated by
distributional semantic models.
Our approach to this problem is to simulate
a brain-based semantic representation proposed by \citeA{Binder16}
using text-based vectors.
This semantic representation comprises 65 attributes based entirely on
functional divisions in the human brain.
Each word is represented as a 65-dimensional vector and each dimension
represents the salience of the corresponding attribute,
namely the degree to which the concept referred to by that word is
related to that attribute.
Because these attributes are based on not only sensorimotor experiences
but also affective, social, and cognitive experiences,
we can analyze distributional word vectors considering
a wide variety of information.
In the analysis, we trained the mapping from the text-based vectors
to the brain-based vectors, by which brain-based vectors of untrained
words are predicted.
Prediction accuracy was measured for each attribute and word
using a leave-one-out cross-validation.

The secondary purpose of this paper is to discuss
the relationship between the embodied theory for abstract words and
distributional semantic models from the results of the analysis.
Recently it has been accepted that 
language or linguistic experience is much more important
for abstract concepts than for concrete concepts,
because abstract words are unlikely to be grounded in perceptual
and sensorimotor experiences, in which concrete concepts are grounded \cite{Borghi17}
A number of approaches have been proposed to explain the role of language
as a simple shortcut \cite{Barsalou08b} or
indirect grounding in perceptual or sensorimotor experiences \cite{Louwerse11,Dove14},
and the need for other information such as
emotional \cite{Kousta11} and social information \cite{Borghi14b}.
The analysis of information encoded in text-based word vectors,
which can be regarded as realizations of linguistic experiences,
is expected to provide some implications for recent embodied
approached to abstract concepts.

\section{Method}
In order to examine what kind of information is encoded in distributional word vectors,
we evaluated how accurately they can simulate \citeS{Binder16} brain-based vectors.
The simulation was performed by training the mapping from text-based vectors
to brain-based vectors and applying the trained mapping to
the text-based vectors of untrained words.
Prediction performance was evaluated by comparing 
the estimated brain-based vectors with the original brain-based vectors.

\begin{table}[t]
  \vspace*{-2mm}
  \caption{Example of words represented as brain-based vectors}
  \label{tbl:word}
  \begin{center}\small
    \renewcommand{\arraystretch}{1.2}
  \begin{tabular}{l@{~~}l|l@{~~~}l} \hline
    Category & \multicolumn{1}{c|}{Word} & Category & \multicolumn{1}{c}{Word} \\ \hline
    plant & apricot, rose, tree & human & actor, girl, parent \hspace*{-2mm}\\
    vehicle & car, subway, boat & social action & celebrate, help\\
    place & airport, lake, lab & visual property & black, new, dark \\ \hline
  \end{tabular}
  \vspace*{-2mm}
  \end{center}
\end{table}

\begin{table}[t]
  \vspace*{-2mm}
  \caption{65 attributes used in brain-based vectors}
  \label{tbl:attribute}
  \begin{center}\small
    \renewcommand{\arraystretch}{1.2}
  \begin{tabular}{l@{~~~}p{65mm}} \hline
    Domain & Attributes \\ \hline
    Vision & Vision, Bright, Dark, Color, Pattern, Large, Small,
             Motion, Biomotion, Fast, Slow, Shape, Complexity, Face, Body \\
    Somatic & Touch, Temperature, Texture, Weight, Pain \\
    Audition & Audition, Loud, Low, High, Sound, Music, Speech \\
    Gustation & Taste \\
    Olfaction & Smell \\
    Motor & Head, UpperLimb, LowerLimb, Practice \\
    Spatial & Landmark, Path, Scene, Near, Toward, Away, Number \\
    Temporal & Time, Duration, Long, Short\\
    Causal & Caused, Consequential\\
    Social & Social, Human, Communication, Self\\
    Cognition & Cognition\\
    Emotion & Benefit, Harm, Pleasant, Unpleasant, Happy, Sad, Angry,
              Disgusted, Fearful, Surprised\\
    Drive & Drive, Needs\\
    Attention & Attention, Arousal\\ \hline
  \end{tabular}
  \vspace*{-2mm}
  \end{center}
\end{table}

\subsection{Brain-based Vectors}
As mentioned above, we used \citeS{Binder16} brain-based componential representation
of words as a gold standard. They provided 65-dimensional vectors of 535 words
comprising 434 nouns, 62 verbs and 39 adjectives, some of which are listed in 
Table\,\ref{tbl:word}.
The dimensions correspond to neurobiologically plausible attributes 
whose neural correlates have been well described.
Table\,\ref{tbl:attribute} lists 65 attributes (and 14 domains)
used in \citeS{Binder16} brain-based vectors.

\subsection{Word Vectors}
In order to ensure the generality of the findings obtained through the analysis,
we constructed six semantic spaces, which were obtained from the combinations of
three distributional semantic models (SGNS, GloVe, PPMI) and
two corpora (COCA and Wikipedia).
As a distributional semantic model, we used three representative models,
namely skip-gram with negative sampling \cite<SGNS;>{Mikolov13a},
GloVe \cite{Pennington14} and positive pointwise mutual information (PPMI)
with SVD \cite{Bullinaria07}.
SGNS and GloVe are prediction-based models that train word vectors by
predicting context words on either side of a target word, 
while PPMI is a counting-based model that trains word vectors by
counting and weighting word occurrences.
We set a vector dimension $d=300$ and a window size $w=10$
for all semantic spaces.

Two corpora used in the analysis were English Wikipedia dump of enwiki-20160601 (Wiki)
and Corpus of Contemporary American English (COCA).
The Wiki and COCA corpora include 1.89G and 0.56G word tokens, respectively.
We built a vocabulary from frequent words that occur 50 times or more
in Wiki corpus
\footnote{%
  Out of 535 words for brain-based vectors, only one word ``joviality''
  was not selected as frequent words for Wiki corpus.
  Hence, we added it to the vocabulary for Wiki corpus.}
or 30 times of more in COCA corpus.
As a result, the vocabulary of Wiki and COCA contained 291,769 words and
108,230 words, respectively.
These two corpora differ in that Wiki is a raw text corpus that is untagged
and unlemmatized, while COCA is a fully tagged and lemmatized corpus.
For Wiki corpus, raw texts were extracted from the dump files
using {\tt WikiExtractor.py}
\footnote{\tt http://medialab.di.unipi.it/wiki/Wikipedia\_Extractor}
and no other preprocessing, such as lemmatization, was applied.

\subsection{Training the Mapping from Text-based Vectors to Brain-based Vectors}
We used two learning methods, namely linear transformation (LT) and
multi-layer perceptron (MLP).
LT trains a mapping matrix $\bmath{M}$ such that $\bmath{B}=\bmath{WM}$ where
$\bmath{B}$ is the matrix with brain-based word vectors as rows
and $\bmath{W}$ is a matrix with text-based word vectors as rows.
MLP trains a neural network with one hidden layer
comprising 150 sigmoid units
and a linear output layer.
In both methods, the mapping was trained
by minimizing the mean squared error,
and gradient descent with AdaGrad was used as an optimization method.

Estimation of brain-based vectors from text-based vectors was performed 
by a leave-one-out cross validation procedure.
For each of the 535 words, we trained the mapping between brain-based and text-based
vectors of the remaining 534 words and estimated a brain-based vector for the target word
using the trained mapping.
By repeating this procedure for all words as a target, we obtained $\widehat{\bmath{B}}$
with estimated brain-based vectors as rows.

\subsection{Performance Measure}
Prediction performance of the estimated vectors was measured
using Spearman's rank correlation $\rho$
between the estimated brain-based matrix $\widehat{\bmath{B}}$ and
the original matrix $\bmath{B}$.\,{\footnotemark}
\footnotetext{%
  Mean squared error can also be a measure for prediction performance.
  However, we are interested in 
  the similarity of order, rather than of absolute value,
  between the original and estimated vectors, and thus we used
  rank correlations in this paper.}
We performed two analyses: column-wise and row-wise matrix correlation.
The column-wise matrix correlation indicates the estimation accuracy for each attribute,
while the row-wise correlation indicates the accuracy for each word.

In addition, we performed a k-means clustering analysis in which 
535 words were grouped into 28 clusters using the estimated brain-based vectors,
and the obtained clustering result was compared with the 28-cluster solution
computed using the original brain-based vectors by \citeA{Binder16}.
The clustering result was evaluated for each gold-standard cluster
by the normalized entropy $H(G_i)$ as follows:
\begin{equation}
  H(G_i) = \frac{-1}{\log|G_i|} \sum_{j=1}^{28} \frac{n_{ij}}{|G_i|} \log{\frac{n_{ij}}{|G_i|}}
\end{equation}
where $G_i$ is the $i$-th gold-standard cluster and $n_{ij}$ denotes
the number of words in $G_i$ that were assigned to the $j$-th estimated cluster.
The normalized entropy represents how diversely words in a word category are
clustered by the estimated vectors. A lower entropy implies that
more words in $G_i$ are grouped into the same cluster.
If and only if all words in $G_i$ are grouped into one cluster, $H(G_i)=0$.

\begin{figure}[t]
  \vspace*{-4mm}
  \includegraphics{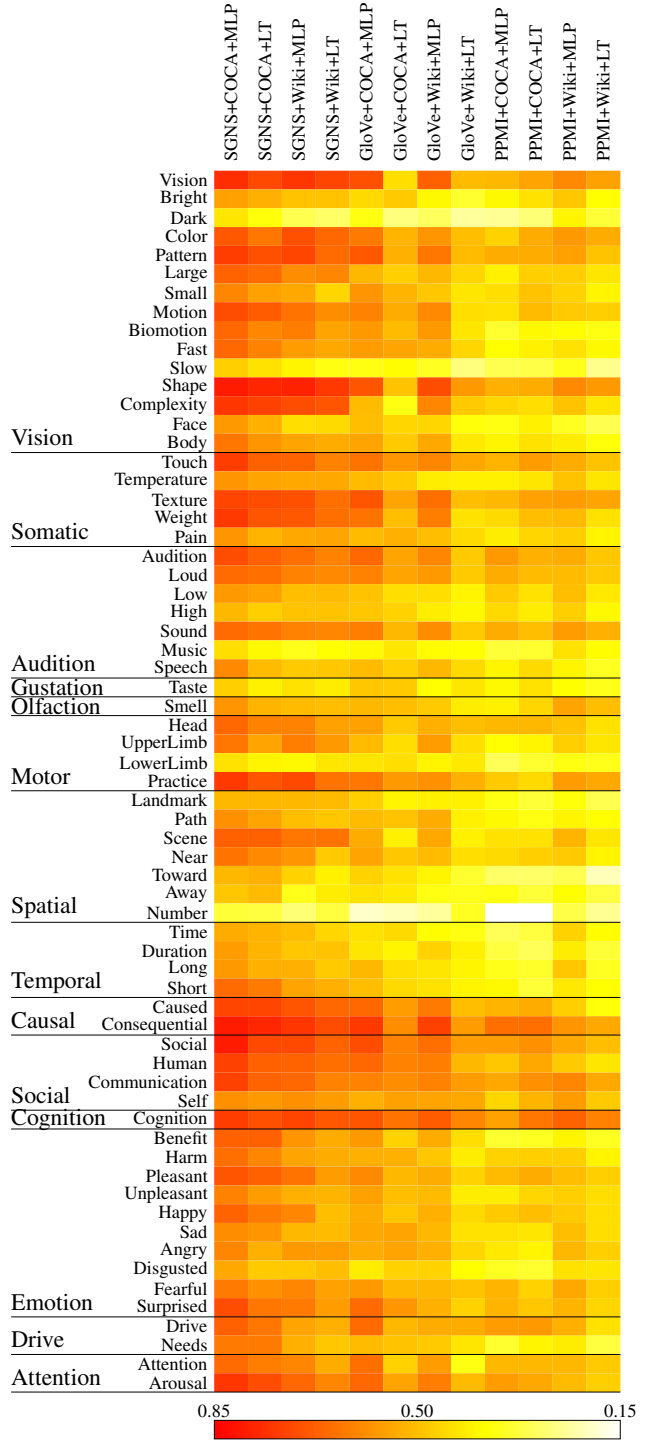}
  \vspace*{-3mm}
  \caption{Correlations between the estimated and original brain-based vectors 
    for 65 attributes. Each row corresponds to the results of an attribute and
    each column shows the results of combinations of
    distributional semantic models (SGNS, GloVe, PPMI), corpora (COCA, Wiki) and
    training methods (MLP, LT).}
  \label{fig:correl-attribute}
  \vspace*{-4mm}
\end{figure}

\begin{figure}[tb]
  \includegraphics{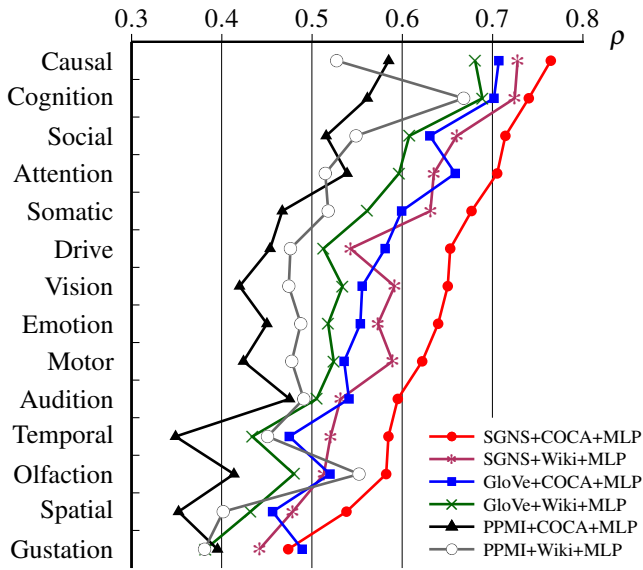}
  \caption{Mean correlations per attribute domain.
    Only the results for MLP are shown for simplicity.}
  \label{fig:correl-domain}
  \vspace*{-2mm}
\end{figure}

\begin{table}[t]
  \vspace*{-2mm}
  \caption{Mean correlations over all attributes}
  \label{tbl:ave-correl}
  \begin{center}
    \renewcommand{\arraystretch}{1.2}
    \begin{tabular}{lllll}\hline
      & & SGNS & GloVe & PPMI \\ \hline
      Wikipedia & MLP & 0.576 & 0.522 & 0.483 \\ 
      & LT & 0.549 & 0.450 & 0.429\\ \hline
      COCA & MLP & 0.634 & 0.554 & 0.440\\ 
      & LT & 0.598 & 0.494 & 0.454 \\ \hline
  \end{tabular}
  \end{center}
  \vspace*{-4mm}
\end{table}

\section{Result}
\subsection{Correlation Analysis by Attribute}
We evaluated the prediction accuracy for attributes
by computing column-wise matrix correlations
between the estimated and original brain-based vector spaces.
Figure\,\ref{fig:correl-attribute} shows correlation
coefficients for 65 attributes.
In addition, these results are summarized in Figure\,\ref{fig:correl-domain},
which depicts mean correlations averaged over attributes of the same domain.

Although in this paper we are not concerned with the overall performance of
word vectors, Table\,\ref{tbl:ave-correl} shows that
SGNS achieved the best prediction performance, and
word vectors trained using the COCA corpus were superior to
those of the Wiki corpus.
In addition, as expected, MLP trained better mappings than LT.
Despite these differences of overall performance,
Figures\,\ref{fig:correl-attribute} and \ref{fig:correl-domain}
demonstrate that relative performance among attributes did not significantly
differ, regardless of distributional model, corpus and training method.

Attributes in {\it causal}, {\it cognitive}, {\it social}, and {\it attentional domains}
were generally predicted with higher accuracy
(i.e., their rank correlations of SGNS+COCA+MLP exceeded 0.7).
In other words, the information of these attributes,
which characterize abstract concepts,
is likely to be encoded in text-based word vectors.
It seems to suggest that
abstract concepts can be largely acquired through linguistic experiences.
On the other hand, sensorimotor and spatiotemporal attributes
were relatively more difficult to predict from text-based word vectors.
This result is consistent with the embodied view of cognition that
perceptual or sensorimotor information for grounding concrete concepts
cannot be acquired through linguistic experiences.
Note that some perceptual attributes such as
{\it vision}, {\it pattern}, {\it shape}, {\it texture} and {\it sound}
were predicted as accurately as abstract attributes, suggesting that  
text-based word vectors can encode these kinds of information.

A somewhat surprising result was that emotional attributes were not
predicted as accurately as social and cognitive ones, although
a large number of NLP studies have demonstrated
successful results of sentiment analysis \cite{Taboada16}.
From a cognitive science (or embodied cognition) perspective, however,
this result suggests that emotional information is more likely to be
acquired from direct emotional experiences than from linguistic ones,
and it is consistent with the view that emotional experiences are
required for grounding abstract concepts \cite{Kousta11,Vigliocco14}.

\begin{figure}[t]
  \vspace*{-4mm}
  \includegraphics{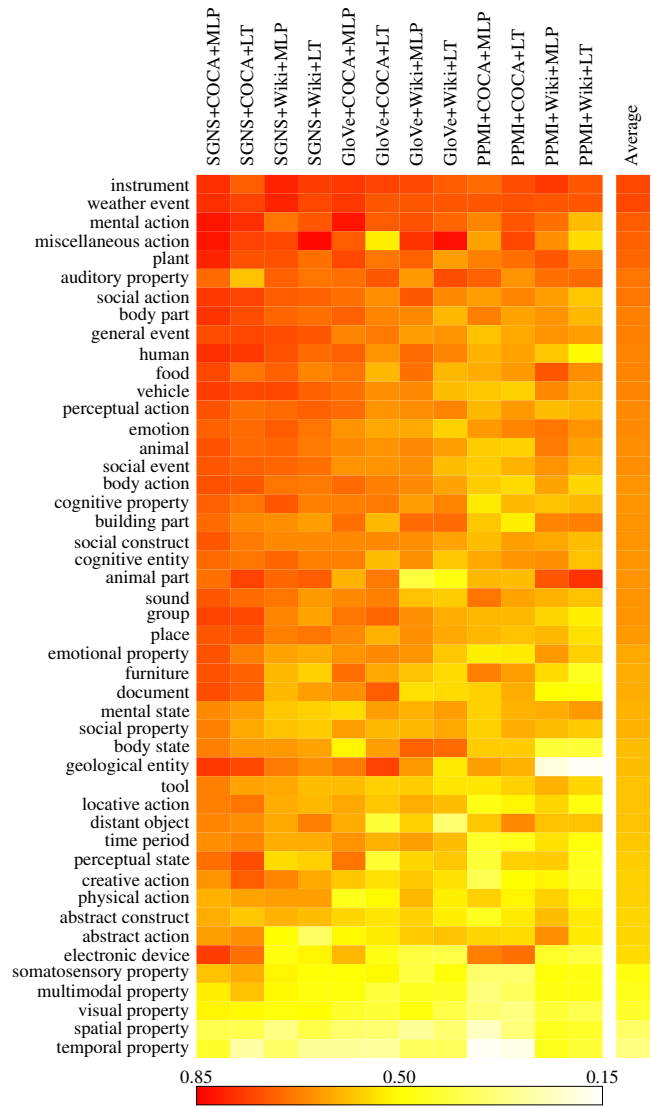}
  \vspace*{-6mm}
  \caption{Mean correlations between the estimated and original brain-based vectors 
    for 47 word categories. Each row corresponds to the results of a word category.}
  \label{fig:correl-word}
  \vspace*{-3mm}
\end{figure}

\subsection{Correlation Analysis by Word}
We computed row-wise matrix correlations between
the estimated and original brain-based vector spaces,
and then averaged these 535 correlations according to 47 word categories.
These word categories are provided a priori by \citeA{Binder16}
and reflect grammatical classes (i.e., noun, verb, adjective) and
semantic classes.\,{\footnotemark}
\footnotetext{%
  Note that word categories provided online slightly differ from those
  shown in \citeS{Binder16} article.
  In this paper, we used the online version of word categories.}
Figure\,\ref{fig:correl-word} shows mean correlations per word category.
As in the case of the attribute analysis,
there were no crucial differences among semantic spaces
and among training methods.

The overall result was that 
brain-based vectors for human-related categories
such as {\it mental action}, {\it social action}, {\it human} and {\it social event}
were relatively better predicted from text-based word vectors.
Emotional and cognitive categories such as {\it emotion} and {\it cognitive property}
were predicted well, but with lower accuracy than human-related categories.
These results are consistent with the findings obtained by the attribute analysis.
On the other hand, other abstract concepts,
in particular many categories of action and property,
were difficult to predict from text-based word vectors.
Distributional semantic models may be insufficient for
representing some kinds of abstract concepts, and
other experiences than linguistic one would be required \cite<e.g.,>{Borghi17}.

Interestingly, many artifact categories such as
{\it instruments}, {\it food}, and {\it vehicle},
and some natural objects such as {\it plant} and {\it animal} 
showed higher prediction performance.
There is no doubt that,
as the embodied theory of language argues, 
these concrete words or concepts are grounded in perceptual and
sensorimotor experiences, but 
some kinds of concrete concepts, in particular artifacts,
may be able to be represented 
(or indirectly grounded) by text-based word vectors.

\begin{figure}[t]
  \includegraphics{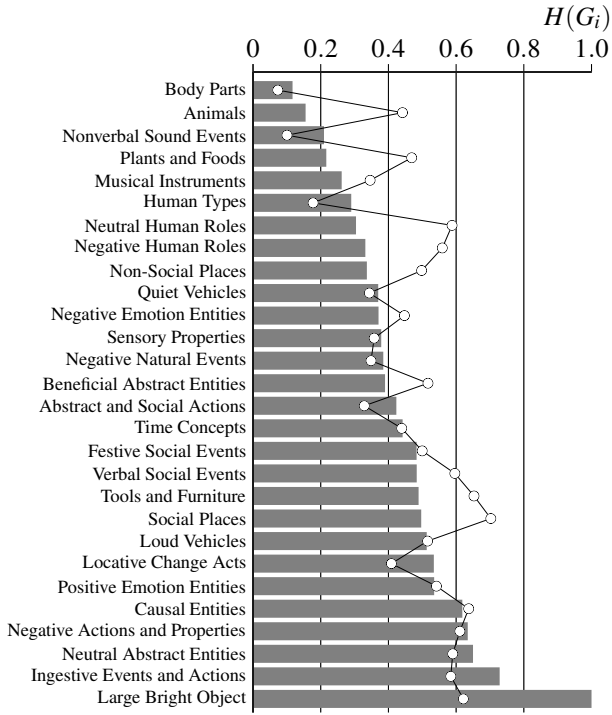}
  \vspace*{-2mm}
  \caption{Normalized entropy of 28 word categories.
    A bar chart represents the result of the estimated brain-based vectors
    for SGNS+COCA+MLP, while a line graph represents the result obtained
    using the original SGNS+COCA vectors.}
  \label{fig:kmeans28-entropy}
  \vspace*{-2mm}
\end{figure}

\subsection{Cluster Analysis}
We performed a cluster analysis in which 535 words were
clustered into 28 clusters by their estimated brain-based vectors.
In the cluster analysis, k-means clustering was used
with k-means++ initialization.
Because a k-means algorithm is nondeterministic
owing to random initialization, we repeated k-means clustering
10 times and averaged mean entropy over these 10 trials.
%
The gold-standard set of 28 word categories is provided through
\citeS{Binder16} data-driven cluster analysis of the original brain-based vectors.
The data-driven clustering revealed several novel distinctions
not considered in the predefined category,
such as the distinction of positive/negative and social/non-social categories.

Figure\,\ref{fig:kmeans28-entropy} shows the result of cluster analysis,
i.e., mean normalized entropies for 28 gold-standard categories,
using the predicted vectors by SGNS+COCA+MLP
(i.e., MLP learning for SGNS+COCA vectors).
The overall mean entropy averaged across categories was 0.434 for
the estimated vectors (SGNS+COCA+MLP)
and 0.459 for the original 300-dimensional SGNS+COCA vectors,
indicating that
some semantic information can be better represented
by mapping the original text-based vectors into
the brain-based vector space.

Figure\,\ref{fig:kmeans28-entropy} demonstrates that
words of human-related categories such as
{\it Human Types}, {\it Neutral Human Roles}, and
{\it Negative Human Roles\/}
were more likely to be grouped into the same clusters
by the estimated vectors, but words in emotional categories
were less likely to belong in the same cluster.
This result is fully consistent with the results of correlation analysis.
In addition, some natural categories such as {\it Animals}
and {\it Plants and Foods}, and artifacts such as
{\it Musical Instruments} and {\it Quiet Vehicles\/}
achieved very low entropy values.
Again, this result suggests the possibility that
some kinds of concrete concepts can be represented by
text-based word vectors without using multimodal information.

\section{Discussion}
In this paper, we have demonstrated that
text-based distributional word vectors can predict
social and cognitive information quite accurately, 
but the accuracy of emotional information is not so high.
Given the existing empirical findings on the importance
of emotion for abstract concepts \cite{Vigliocco14,Buccino16},
this result suggests that 
direct emotional experiences are necessary for grounding
abstract concepts, and thus may lend support to 
some embodied theories \cite{Kousta11,Vigliocco14}.
On the other hand, some other embodied theories
such as WAT theory \cite{Borghi14b} have argued that
social experiences also play an important role in
representation of abstract concepts.
However, the result of our analysis that 
social information can be conveyed by language 
may diminish the importance of social experiences for concrete concepts.
Furthermore, the need of social-cognitive ability is not specific to 
abstract concepts;
concrete concepts are acquired and processed through
social abilities such as a Theory of Mind \cite<e.g.,>{Bloom00}.

It was also found from the analysis that 
perceptual, sensorimotor and spatiotemporal information is
less likely to be encoded in word vectors. 
This is what is expected from a number of studies 
claiming that distributional semantic models learn only from
co-occurrences of amodal symbols that are not grounded
in the real world \cite{Glenberg00}.
It is also consistent with the findings of multimodal
distributional semantics that 
inclusion of visual information improves
semantic representation for concrete words \cite<e.g.,>{Kiela14}.
At the same time, the analysis also suggested
the possibility that 
some perceptual information can be derived
from distributional semantic models.
This result does not deny the embodied account that
grounding in perceptual and sensorimotor experiences is necessary
for representing and acquiring concrete concepts.
For practical applications to NLP and AI, however,
text-based word vectors can possibly
provide enough information without considering
the embodied nature of word meanings.

Of course, the analysis presented in this paper is not comprehensive
and has some limitations.
One important limitation is that the brain-based vectors
represent the salience of attributes that characterize concepts,
but do not necessarily represent the value of salient attributes.
For some attributes such as {\it Bright} and {\it Happy},
their value is indistinguishable from their salience, but 
many other attributes such as {\it Color} and {\it Human}
have distinct values independent of their salience.
Hence, the analysis in this paper cannot examine
the representational power of attribute values.
Our analysis is also limited within a small set of vocabulary words.
To generalize and refine the findings presented in this paper,
we have to evaluate a much larger set of vocabulary words that
are not included in \citeS{Binder16} dataset.
It would be interesting and vital for further work
to extend the analysis and to develop a novel analysis method
so as to overcome these limitations.

\section{Acknowledgments}
This research was supported by JSPS KAKENHI Grant Numbers JP15H02713
and SCAT Research Grant.

\bibleftmargin=3ex
\bibindent=-3ex

\end{document}